\documentclass[twoside,leqno,twocolumn]{article}

\usepackage[letterpaper]{geometry}

\usepackage{siamproceedings}
\usepackage{chngcntr}
\counterwithout{figure}{section}
\counterwithout{table}{section}
\counterwithout{algorithm}{section}
\usepackage[T1]{fontenc}
\usepackage{amsfonts}
\usepackage{graphicx}
\usepackage{epstopdf}
\usepackage{enumitem}
\usepackage{algorithmic}
\ifpdf
  \DeclareGraphicsExtensions{.eps,.pdf,.png,.jpg}
\else
  \DeclareGraphicsExtensions{.eps}
\fi

\newsiamremark{remark}{Remark}
\newsiamremark{hypothesis}{Hypothesis}
\crefname{hypothesis}{Hypothesis}{Hypotheses}
\newsiamthm{claim}{Claim}

\usepackage{amsopn}

\usepackage{tikz}
\usetikzlibrary{arrows.meta, positioning, calc}
\usepackage{pifont}
\usepackage{booktabs}
\usepackage{multirow}
\usepackage{array}
\usepackage[normalem]{ulem}          
\usepackage[table]{xcolor}
\newcommand{\spm}[1]{{\scriptsize$\pm$#1}}
\definecolor{categorycolor}{RGB}{195, 238, 236}

\begin{document}

\newcommand\relatedversion{}
\renewcommand\relatedversion{\thanks{The full version of the paper can be accessed at \protect\url{https://arxiv.org/abs/0000.00000}}} 

\title{\Large Artemis: Anatomy-Resolved inTervention for Eliminating Multimodal NeuroImage confounderS}
\author{Siyuan Dai \footnotemark[1]
\and Yang Du \footnotemark[1]
\and Kun Zhao \footnotemark[1]
\and Zhusuyi Chen \thanks{Department of Electrical and Computer Engineering, University of Pittsburgh, Pittsburgh, PA. \email{siyuan.dai@pitt.edu} \email{liang.zhan@pitt.edu}}
\and Heng Huang \thanks{Department of Computer Science, University of Maryland, College Park, MD}
\and Paul Thompson \thanks{Imaging Genetics Center, Mark \& Mary Stevens Institute for Neuroimaging \& Informatics, Keck School of Medicine, University of Southern California, Los Angeles, CA}
\and Chao Shi \thanks{School of Systems Science and Industrial Engineering, Binghamton University, Binghamton, NY}
\and Haoteng Tang \thanks{Department of Computer Science, University of Texas Rio Grande Valley, Edinburg, TX}
\and Liang Zhan\footnotemark[1] 
}
\date{}

\maketitle
\fancyfoot[R]{\scriptsize{Copyright \textcopyright\ 20XX by SIAM\\
Unauthorized reproduction of this article is prohibited}}





\begin{abstract}
Multimodal neuroimaging, integrating functional connectivity from fMRI and structural connectivity from DTI, enables non-invasive analysis of brain networks using graph neural networks. However, demographic factors such as age and sex systematically confound the relationship between brain connectivity and clinical outcomes, causing GNNs to exploit spurious shortcuts rather than learning causally invariant representations. While recent causal GNN methods introduce causality at the graph-modeling level, their causal mechanisms remain domain-agnostic without accounting for the real-world confounders inherent in clinical neuroimaging data. Moreover, brain networks are constructed from atlas-based parcellations where each region exhibits distinct sensitivity to demographic factors, necessitating region-aware adjustment. We propose Artemis, a region-level causal framework that bridges this gap with causal intervention at each brain region independently by learning region-specific confounder representations 
with lightweight parameters. Our adjustment comprehensively utilized the multimodal functional and structural features for graph reasoning as a plug-in module compatible with arbitrary GNN backbones. Experiments on three benchmarks, ADNI for disease diagnosis, OASIS for dementia staging, and HCP for sex classification, demonstrate consistent improvements over representative GNN-based baselines.
Multiple supporting experiments further demonstrate statistical significance and neuroscientific interpretability.
\end{abstract}

\section{Introduction.}
\label{sec:intro}

\begin{figure*}[!t]
    \centering
    \includegraphics[width=0.9\textwidth]{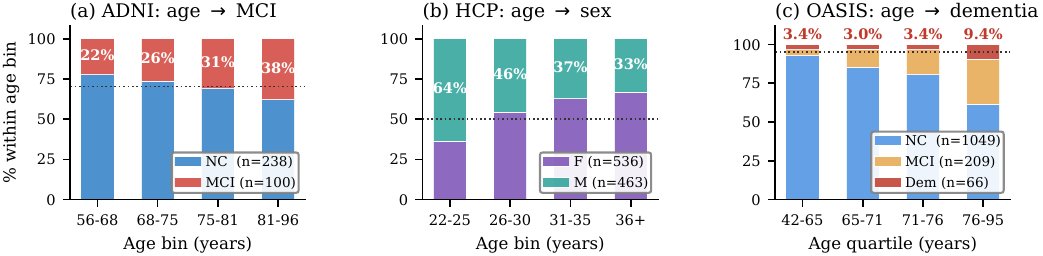}
    \caption{%
        Demographic confounders are entangled with labels across all three benchmarks.
        (a)~On ADNI, MCI prevalence climbs from the youngest age to oldest.
        (b)~On HCP, the younger cohort is male-skewed.
        (c)~On OASIS, dementia prevalence roughly triples in the oldest age quartile relative to the three younger ones.
    }
    \label{fig:teaser}
\end{figure*}

Non-invasive neuroimaging is a cornerstone of clinical neuroscience.
Functional MRI (fMRI) captures the temporal co-activation of brain regions, while diffusion tensor imaging (DTI) delineates the anatomical white-matter pathways that support this activity.
Modeling each modality as a graph over atlas-defined regions of interest (ROIs) lets graph neural networks (GNNs) learn discriminative representations for disease diagnosis, cognitive prediction, and demographic analysis~\cite{Li2020BrainGNNIB, Kawahara2017BrainNetCNNCN, Kan2022BrainNT}.
Fusing the two modalities further provides a richer, anatomically grounded substrate for population-level brain analysis~\cite{Tang2024InterpretableSE, Yin2024AHG, Ye2023BidirectionalMW}.

However, demographic confounding in multimodal clinical neuroimages is pervasive and largely overlooked in downstream predictive modeling.
Clinical cohorts are naturally demographically unbalanced: age, sex, and education co-vary with nearly every clinical label of interest.
For instance, in ADNI \cite{Jack2008TheAD}, subjects with mild cognitive impairment (MCI) are on average older than healthy controls, as demonstrated in Fig.~\ref{fig:teaser}, while in HCP~\cite{Essen2013TheWH}, sex is associated with educational differences.
Meanwhile, demographic factors are known to shape regional brain structure and function, including age-related decline in hippocampal integrity and sex-related effects in cortical regions such as the primary visual cortex~\cite{Ritchie2017SexDI,Fjell2014WhatIN}.
Without explicit adjustment, GNNs trained end-to-end can therefore rely on demographic-driven connectivity patterns as spurious shortcuts, neglecting causally relevant disease signals.
This issue can lead to biased predictions, degraded robustness across subpopulations, and misleading neuroscientific interpretations, all of which are especially undesirable in clinical applications.

Existing causal GNNs, while promising, remain domain-agnostic.
Typical works inject causal reasoning into GNNs, most prominently by disentangling \emph{"causal"} and \emph{"spurious"} subgraphs via intervention or invariance objectives~\cite{Fan2022DebiasingGN, Sui2021CausalAF, Chen2022LearningCI}.
These methods advance causally robust graph learning in general, but their notion of a confounder is \emph{structural}: induced by the graph topology itself rather than by any external variable.
For neuroimages, CI-GNN~\cite{zheng2024ci} employs Granger causality as a post-hoc interpretability tool, and Contrasformer~\cite{xu2024contrasformer} addresses sub-population shift via contrast graphs without invoking causal adjustment at all.
Such approaches implicitly assume that confounding lives inside the connectivity matrix, which is unobservable, ignoring the fact that in clinical cohorts, the dominant confounders are \emph{observed} demographic attributes with well-documented neuroscientific effects.
As a consequence, current causal GNNs cannot perform the most basic causal operation that neuroimages necessitate: adjusting for known demographic confounders via backdoor adjustment.

Furthermore, demographic confounding is not uniform across the brain \cite{Alex2023AGM,Eickhoff2018ImagingbasedPO}.
Atlas-based parcellations divide the cortex and subcortex into anatomically and functionally distinct regions, each with its own developmental trajectory and sensitivity profile.
A single global confounder correction applied uniformly to all ROIs cannot capture this heterogeneity, and risks either under-correcting regions like the hippocampus or over-correcting regions that are already clean, which are strongly affected by both aging and Alzheimer's pathology \cite{Frisoni1999HippocampalAE}.

To address these gaps, we introduce Artemis, a region-level causal intervention framework motivated by backdoor adjustment and illustrated in Fig.~\ref{fig:teaser}.
Artemis maps the demographics vector to a \emph{per-ROI} confounder embedding through a shared multilayer perceptron combined with learnable region tokens, capturing region-specific sensitivity.
We introduce a lightweight exponential-moving-average (EMA) memory bank that maintains a running per-ROI estimate of the population confounder distribution, enabling each sample's confounder to be centered against the cohort mean, a low-variance approximation of the backdoor-adjustment expectation.
The entire intervention adds only a few thousand parameters and plugs into any GNN backbone, making it a drop-in module rather than a new architecture.
Across three clinical benchmarks, ADNI (NC vs. MCI), HCP (sex), and OASIS (CDR three-class), our Artemis outperforms ten representative GNN-based baselines across multiple categories, improving accuracy over the vanilla GCN backbone by $+20.9\%$, $+27.9\%$, and $+7.8\%$, and AUC by $+26.2\%$, $+34.2\%$, and $+8.0\%$, respectively.
We summarize our contributions as follows.
\begin{itemize}[leftmargin=*, nosep]
    \item We identify region-specific demographic confounding as an overlooked but crucial source of spurious shortcuts in multimodal brain-network GNNs.
    \item We propose \textsc{Artemis} by formulating a region-level backdoor adjustment, a lightweight plug-in intervention module with only $7K$ parameters, which is compatible with arbitrary GNN backbones.
    \item On three clinical benchmarks, Artemis consistently outperforms ten representative GNN-based baselines across multiple categories, with substantial improvements in accuracy, F1, and AUC. Multiple supportive analyses validate the formulated region-level backdoor adjustment.
\end{itemize}

\section{Related Works.}
\label{sec:related}

\begin{figure*}[!t]
    \centering
    \includegraphics[width=\textwidth]{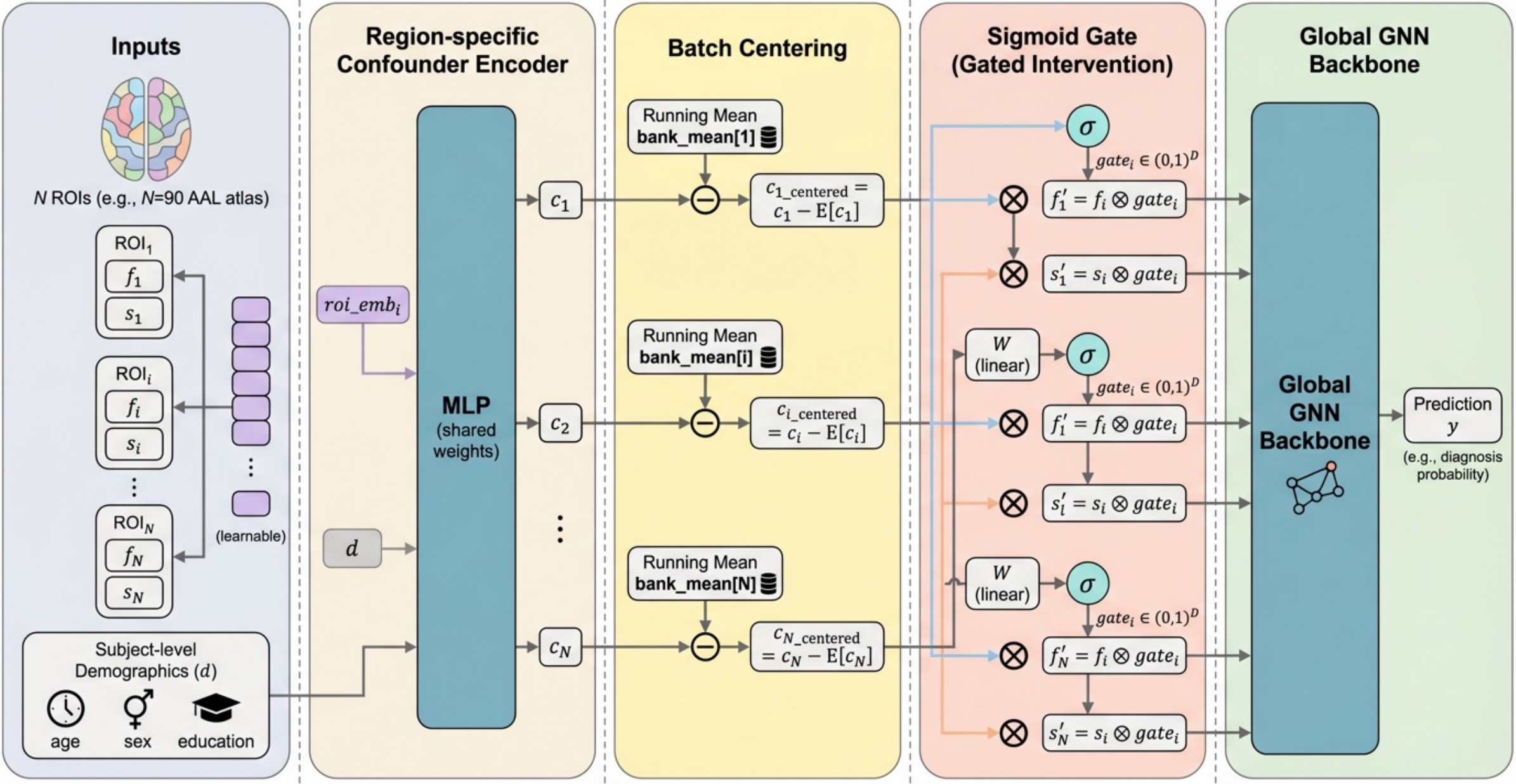}
    \caption{Artemis pipeline. (1) Per-ROI multimodal features together with subject-level demographics $d$ serve as inputs. (2) A shared MLP combined with learnable per-ROI tokens $\mathtt{roi\_emb}_i$ produces region-specific confounder embeddings $c_i$. (3) A per-ROI EMA memory bank stores the running mean of $c_i$ over the training population, approximating $\mathbb{E}[c_i]$ for backdoor centering. (4) A learned gate $\sigma(W c_i^{\mathrm{centered}})$ is applied elementwise ($\odot$) to \emph{both} $f_i$ and $s_i$. (5) The adjusted features feed any GNN backbone for downstream prediction.}
    \label{fig:method}
\end{figure*}

\subsection{Brain-Network GNNs.}
Graph neural networks have become the dominant paradigm for learning over brain connectomes.
BrainNetCNN~\cite{Kawahara2017BrainNetCNNCN} pioneered the use of edge-to-edge and edge-to-node convolutions tailored to symmetric connectivity, and BrainGNN~\cite{Li2020BrainGNNIB} introduced ROI-aware graph convolutions with a top-$K$ pooling mechanism that highlights clinically salient regions.
More recently, transformer-style architectures have been adapted to brain networks: BrainNetTF~\cite{Kan2022BrainNT} employs a cluster-readout module that captures community structure in functional connectivity, and BioBGT~\cite{Peng2025BiologicallyPB} incorporates spectral positional encodings and community-guided attention to inject biological priors into the attention mechanism.
Multimodal extensions exploit the complementarity of functional and structural connectivity: Tang et al.~\cite{Tang2024InterpretableSE} propose an interpretable FC-SC fusion framework for Alzheimer's disease staging, and Yin et al.~\cite{Yin2024AHG} align multi-view connectivity through heterogeneous graph attention.
Despite architectural diversity, these models are trained end-to-end on class labels alone and never account for demographic variables that co-vary with both the input graphs and the clinical target, leading to exploit demographic shortcuts for only reducing the training loss.

\subsection{Causal Reasoning on Graphs}
One line of work injects causal reasoning into general-purpose GNNs, typically by disentangling a causal subgraph from a spurious complement:
DIR~\cite{wu2022discovering} partitions each input into invariant and variant components.
CAL~\cite{Sui2021CausalAF} implements this via causal attention and do-calculus-style training,
and GIL~\cite{li2022learning}, CIGA~\cite{Chen2022LearningCI}, GSAT~\cite{miao2022interpretable}, MoleOOD~\cite{yang2022learning}, and Fan~et~al.~\cite{Fan2022DebiasingGN} extend this agenda through graph-level invariance, sparse stochastic attention, environment-invariant features, or stable-learning objectives.
A separate thread brings causality specifically to brain networks: CI-GNN~\cite{zheng2024ci} uses Granger-causal interactions as a \emph{post-hoc} interpretability tool without any adjustment during learning. While Contrasformer~\cite{xu2024contrasformer} targets sub-population shift via a contrast graph that is distributional rather than causal in the do-calculus sense, and MediAD~\cite{jin2025cross} pursues a heavyweight, LLM-augmented cross-modal causal view at the patient level.
A growing literature on algorithmic fairness in medical imaging further highlights that demographic attributes routinely induce subgroup disparities~\cite{seyyed2021underdiagnosis, petersen2023path}, motivating adjustment without providing a graph-level causal mechanism.
Across both threads, the confounder is either treated as \emph{latent graph structure} or handled only distributionally, none of these methods adjusts for the \emph{observed} demographic variables that are known a priori and dominate confounding in clinical brain-network studies.

\section{Methodology.}
\label{sec:method}

As shown in Fig~\ref{fig:method}, we summarize the proposed Artemis pipeline: a region-specific confounder encoder, an EMA memory bank, and a gated multimodal intervention plug into any GNN backbone. The following subsections formalize each component.

\subsection{Problem Formulation and Causal Graph.}

\begin{figure}[t]
    \centering
    \includegraphics[width=0.9\columnwidth]{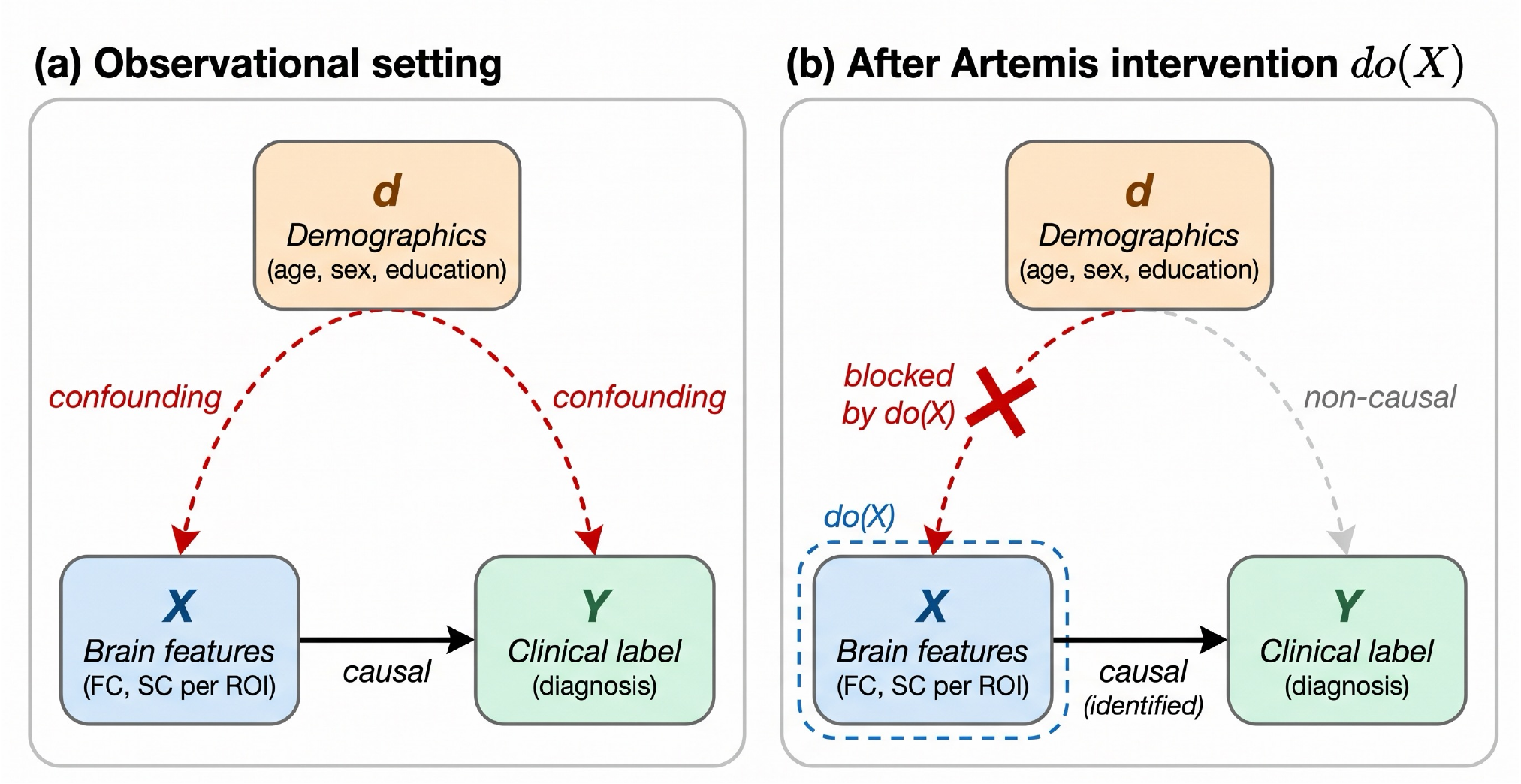}
    \caption{Causal graph for brain-network classification.
    (a)~Observed demographics $d$ confound both brain features $X$ and the clinical label $Y$ via backdoor paths.
    (b)~Artemis performs backdoor adjustment by intervening on the $d\!\to\!X$ path at each ROI.}
    \label{fig:dag}
\end{figure}

For each subject, we are given a multimodal brain network defined over a fixed atlas of $N$ regions of interest (ROIs).
The functional connectivity matrix $\mathrm{FC}\in\mathbb{R}^{N\times N}$ is computed from resting-state fMRI and the structural connectivity matrix $\mathrm{SC}\in\mathbb{R}^{N\times N}$ from DTI tractography.
We treat each $ROI_i$ as a node whose multimodal feature is the concatenation of its FC and SC rows, written as $f_i\in\mathbb{R}^{N}$ and $s_i\in\mathbb{R}^{N}$, so that the node-feature tensor is $X=\{(f_i,s_i)\}_{i=1}^{N}$.
In addition, each subject carries an observed demographics vector $d\in\mathbb{R}^{d_{\mathrm{demo}}}$ comprising age, sex, and education (actual involved attributes depend on the dataset), and a prediction target $y$.

Following the most common backdoor path in neuroimage analysis domain (e.g.\ \emph{demographics} $\rightarrow$ \emph{connectivity} $\rightarrow$ \emph{prediction}), we adopt the causal directed acyclic graph illustrated in Fig~\ref{fig:dag}: $d\!\to\!X$, $d\!\to\!y$, $X\!\to\!y$, in which $d$ is an observed common cause, demographics of both the imaging features and the clinical label.
Standard maximum-likelihood training of $P(y\mid X)$ leaves the backdoor path $X\!\leftarrow\!d\!\to\!y$ open and is therefore prone to demographic shortcuts.
Our objective is the interventional distribution $P(y\mid \mathrm{do}(X))$, which by the backdoor criterion~\cite{pearl2009causality} admits the identification
\begin{equation}
P(y\mid \mathrm{do}(X)) \;=\; \mathbb{E}_{d}\!\left[\,P(y\mid X, d)\,\right].
\label{eq:backdoor}
\end{equation}
Crucially, Eq.~\eqref{eq:backdoor} adjusts at the \emph{feature} level rather than at the label level, it requires modeling how $d$ modulates the per-ROI features and then marginalizing over the population distribution of $d$.

\subsection{Region-Specific Confounder Encoder.}

Different brain regions exhibit markedly different sensitivity to demographic factors, e.g., the hippocampus and entorhinal cortex are dominated by age-related atrophy \cite{Alex2023AGM, Eickhoff2018ImagingbasedPO}.
A single global confounder vector applied uniformly to all ROIs cannot respect this heterogeneity.
At the same time, instantiating $N$ independent MLPs is parameter-inefficient and discards inductive bias across regions.

We therefore implement \emph{region specificity} via a single shared MLP combined with a learnable per-ROI token.
Let $\mathtt{roi\_emb}\in\mathbb{R}^{N\times d_{\mathrm{roi}}}$ be a learnable embedding table whose $i$-th row $\mathtt{roi\_emb}_i$ is a small, randomly initialized identity vector for $ROI_i$.
The per-ROI confounder embedding is
\begin{equation}
c_i \;=\; \mathrm{MLP}\!\left(\,[\,d\,;\,\mathtt{roi\_emb}_i\,]\,\right) \;\in\;\mathbb{R}^{d_c},
\end{equation}
which can be viewed as a feature-wise conditioning of the demographics vector by an ROI identity~\cite{perez2018film}.
Although the same MLP weights are used for every region, the ROI token shifts its input and yields an ROI-specific output, so identical demographics produce different $c_i$ for different regions.
The total parameter count of the encoder is $|\mathrm{MLP}| + N\!\cdot\!d_{\mathrm{roi}}$, which is two orders of magnitude smaller than $N$ separate MLPs (e.g., $90\!\times\!16=1440$ extra parameters for ADNI with $d_{\mathrm{roi}}=16$).
Crucially, $c_i$ is not a $[0,1]$ saliency or an importance weight, it is an unbounded embedding that summarizes how this region responds along the demographics axes, i.e., a region-conditional surrogate for $P(c\mid d, \mathrm{ROI}=i)$.

\subsection{Population-Level Memory Bank and Backdoor Centering.}

The backdoor adjustment in Eq.~\eqref{eq:backdoor} requires the population-level expectation $\mathbb{E}_d[c_i]$ for every ROI.
Computing this exactly at every step is infeasible under mini-batch training, and per-batch means are noisy on small clinical cohorts.
We therefore maintain a per-ROI memory bank $B\in\mathbb{R}^{N\times d_c}$, registered as a non-trainable buffer and updated by an exponential moving average (EMA):
\begin{equation}
B_i \;\leftarrow\; m\cdot B_i \;+\; (1-m)\cdot \tfrac{1}{|\mathcal{B}|}\!\sum_{j\in\mathcal{B}} c_i^{(j)},
\label{eq:ema}
\end{equation}
where $\mathcal{B}$ is the current mini-batch and $m\in[0.9,0.999]$ is the momentum (default $m=0.999$).
The first batch initializes $B_i$ directly to its batch mean rather than via Eq.~\eqref{eq:ema}, avoiding the cold-start bias of starting from zero.

The bank is then used to \emph{center} the confounder embedding before any downstream use:
\begin{equation}
c_i^{\mathrm{centered}} \;=\; c_i \;-\; B_i.
\end{equation}
Two points are worth emphasizing.
First, $B$ is a \emph{consistent, low-variance approximation} of $\mathbb{E}_d[c_i]$ rather than a strictly unbiased Monte-Carlo estimate: the EMA weights do not sum to one and the underlying encoder drifts during training, so $B$ averages snapshots of a non-stationary distribution.
Once training has converged, however, the effective sample size of order $1/(1-m)\!\approx\!10^{3}$ drives the variance well below that of any single mini-batch mean.
Significantly, the bank is frozen at inference time, exactly as the running statistics of batch normalization are frozen at evaluation~\cite{ioffe2015batch}.
This guarantees train-test consistency that every test subject is centered against the same population reference that the model was calibrated to during training.

\subsection{Gated Multimodal Intervention}

\begin{figure}[t]
    \centering
    \includegraphics[width=\columnwidth]{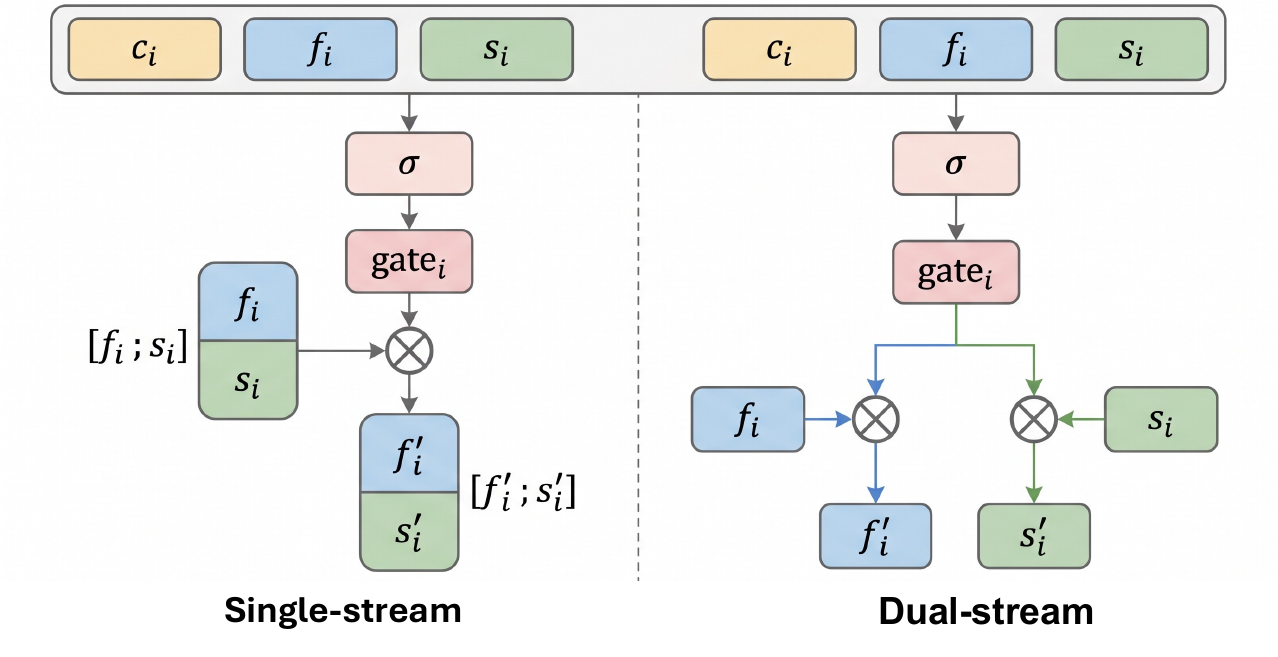}
    \caption{Single-stream vs. dual-stream gated intervention for one ROI.
    Left: a single gate is applied to the concatenated multimodal feature $[f_i; s_i]$.
    Right (Artemis): the same gate is applied independently to the $f_i$ and $s_i$ streams.}
    \label{fig:intervention}
\end{figure}

The centered confounder $c_i^{\mathrm{centered}}$ drives a learned gate that selectively suppresses confounded directions in the per-ROI multimodal features.
Let $W\in\mathbb{R}^{d_{\mathrm{feat}}\times d_c}$ and $b\in\mathbb{R}^{d_{\mathrm{feat}}}$ be a single linear layer, where $d_{\mathrm{feat}}$ is the projected feature dimension shared by the FC and SC streams.
The gate is
\begin{equation}
g_i \;=\; \sigma\!\left(W\,c_i^{\mathrm{centered}} + b\right) \;\in\;(0,1)^{d_{\mathrm{feat}}},
\end{equation}
and is applied elementwise to \emph{both} modalities of region $i$:
\begin{equation}
f_i' \;=\; f_i \odot g_i, \qquad s_i' \;=\; s_i \odot g_i.
\end{equation}
We stress that $g_i$ is \emph{not} attention.
It is a per-ROI scalar mask whose drive signal is the demographic confounder rather than pairwise ROI similarity.
Intuitively, $|g_{i,k}-0.5|$ encodes how much confounder contamination feature dimension $k$ of region $i$ carries, given this subject's demographics, and the linear layer is shaped by the downstream task loss to push the mask towards $0$ along precisely those directions whose variance is explained by $d$.
Gate strength is therefore an \emph{interaction} between the demographic sensitivity of $c_i$ and the task-driven suppression learned by $W$.

The adjusted features $\{(f_i', s_i')\}_{i=1}^{N}$, together with edges derived from the SC matrix, are passed to any GNN backbone.
We use a 2-layer GCN by default, but the intervention is architecture-agnostic.
In this case, the total parameter overhead of the intervention, encoder MLP, ROI embedding table, EMA buffer, and gate layer, is only on the order of a few thousand parameters (about $7$k in our default configuration), which is negligible compared with any modern GNN-based backbone.

\section{Experiments.}
\label{sec:experiments}

\begin{figure*}[!t]
    \centering
    \includegraphics[width=0.9\textwidth]{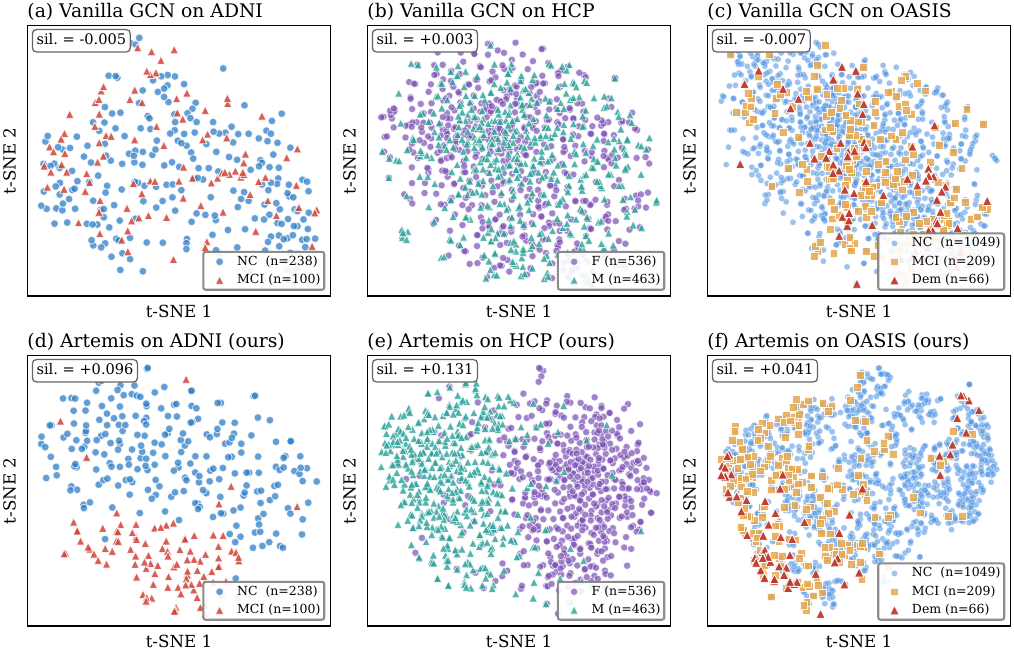}
    \caption{%
        Pre-classifier embedding visualization via t-SNE on all three benchmarks (silhouette scores, raw $d$-dim embedding wrt task label).%
    }
    \label{fig:tsne_support}
\end{figure*}

\begin{figure}[t]
    \centering
    \includegraphics[width=0.9\columnwidth]{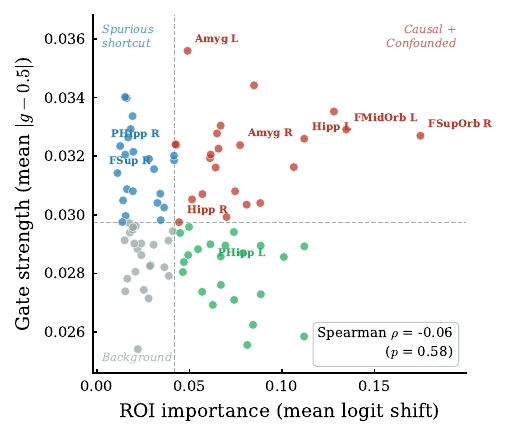}
    \caption{Gate strength vs.\ permutation importance for 90 ADNI ROIs (AAL atlas). The two quantities are largely uncorrelated (Spearman $\rho = -0.06$, $p = 0.58$). Upper-right: clinically relevant regions that are also demographically confounded (e.g., hippocampus). Upper-left: age-sensitive but clinically irrelevant regions.}
    \label{fig:quadrant}
\end{figure}

\subsection{Experimental Setup.}
We evaluate Artemis on three clinical neuroimaging benchmarks.
\textbf{ADNI} (NC vs.\ MCI, $N\!=\!90$ AAL ROIs, $n\!=\!338$) uses age and sex as observed confounders.
\textbf{HCP} (sex classification, $N\!=\!82$ Desikan-Killiany ROIs, $n\!=\!999$) uses age and education.
\textbf{OASIS} (CDR three-class: NC/MCI/Dementia, $N\!=\!132$ Harvard-Oxford ROIs, $n\!=\!1324$) uses age and sex as confounders.

We compare against eleven baselines organized in three groups: \emph{general GNN baselines} (GCN \cite{kipf2016semi}, GAT \cite{velivckovic2018graph}, GIN \cite{xu2018powerful}), \emph{brain-specific architectures} (BrainGNN~\cite{Li2020BrainGNNIB}, BrainNetTF~\cite{Kan2022BrainNT}, BrainNetCNN~\cite{Kawahara2017BrainNetCNNCN}, BioBGT), and \emph{causality-related GNNs} (CI-GNN~\cite{zheng2024ci}, Contrasformer~\cite{xu2024contrasformer}, CAL~\cite{Sui2021CausalAF}).

Unless stated otherwise, Artemis defaultly uses $d_c\!=\!32$, $d_{\mathrm{roi}}\!=\!16$, $d_{\mathrm{feat}}\!=\!64$, a 2-layer GCN backbone, and EMA momentum $m\!=\!0.999$.
The training objective is a single cross-entropy loss without introducing any auxiliary loss.
All models are evaluated with fixed 5-fold subject-level cross-validation, with early stopping on validation accuracy (patience $20$) and a maximum of $100$ epochs. We report accuracy, macro-F1, and macro-averaged ROC-AUC (one-vs-rest for the three-class OASIS task) as our evaluation metrics.

\subsection{Main Results}


\begin{table*}[t]
\centering
\caption{Experimental results on ADNI, HCP, and OASIS
         (5-fold cross-validation, mean\,$\pm$\,std, \%).
         Acc = accuracy; F1 = macro-F1; AUC = macro-averaged ROC-AUC
         (one-vs-rest for the 3-class OASIS task).
         \textbf{Bold} = best overall;
         \underline{underline} = second best.
         }
\label{tab:results}
\footnotesize
\setlength{\tabcolsep}{3.0pt}
\renewcommand{\arraystretch}{1.05}
\begin{tabular}{l l r ccc ccc ccc}
\toprule
& & & \multicolumn{3}{c}{\textbf{ADNI} (NC/MCI, $n$=338)}
  & \multicolumn{3}{c}{\textbf{HCP} (Gender, $n$=999)}
  & \multicolumn{3}{c}{\textbf{OASIS} (CDR 3-cls, $n$=1324)} \\
\cmidrule(lr){4-6}\cmidrule(lr){7-9}\cmidrule(lr){10-12}
& \textbf{Method} & \textbf{Params}
  & Acc & F1 & AUC
  & Acc & F1 & AUC
  & Acc & F1 & AUC \\
\midrule

\rowcolor{categorycolor}
\multicolumn{12}{l}{\textit{\textbf{General GNNs}}} \\
& GCN & 12K
  & 63.4\spm{13.2} & 46.2\spm{11.6} & 59.2\spm{10.1}
  & 58.8\spm{5.0}  & 56.7\spm{7.8}  & 58.2\spm{6.3}
  & 50.1\spm{5.5}  & 40.4\spm{13.6} & 66.8\spm{6.0}  \\
& GAT & 13K
  & 67.2\spm{5.4}  & 41.2\spm{8.2}  & 63.1\spm{5.6}
  & 57.8\spm{1.2}  & 54.3\spm{4.5}  & 60.3\spm{3.8}
  & 46.8\spm{4.0}  & 41.0\spm{3.2}  & 67.3\spm{3.9}  \\
& GIN & 21K
  & 70.1\spm{3.4}  & 36.2\spm{12.7} & 59.9\spm{2.1}
  & 67.4\spm{2.0}  & 64.3\spm{3.5}  & 70.8\spm{3.1}
  & 49.7\spm{2.8}  & 36.8\spm{2.3}  & 65.0\spm{2.4}  \\

\rowcolor{categorycolor}
\multicolumn{12}{l}{\textit{\textbf{Brain-specific}}} \\
& BrainGNN & 46K
  & 61.2\spm{5.4}  & 45.0\spm{9.0}  & 62.3\spm{6.3}
  & 58.3\spm{2.0}  & 52.1\spm{8.8}  & 58.3\spm{0.8}
  & 42.7\spm{5.7}  & 32.1\spm{7.7}  & 56.7\spm{7.0}  \\
& BrainNetTF & 164K
  & 76.3\spm{3.3}  & 47.4\spm{9.8}  & 65.6\spm{6.9}
  & 60.3\spm{2.3}  & 56.9\spm{5.1}  & 60.3\spm{1.5}
  & 52.0\spm{3.2}  & 45.5\spm{3.7}  & 70.0\spm{3.5}  \\
& BrainNetCNN & 882K
  & \underline{79.3\spm{5.3}}  & \underline{52.4\spm{26.8}} & \underline{76.0\spm{13.4}}
  & \underline{80.5\spm{13.8}} & \underline{68.6\spm{34.6}} & \underline{83.7\spm{17.1}}
  & 48.5\spm{7.6}  & 43.2\spm{7.2}  & 69.6\spm{9.9}  \\
& BioBGT & 83K
  & 70.4\spm{4.9}  & 48.1\spm{5.9}  & 67.5\spm{5.8}
  & 61.7\spm{1.3}  & 58.3\spm{3.1}  & 62.3\spm{1.1}
  & 48.7\spm{4.8}  & 37.4\spm{4.8}  & 66.6\spm{3.8}  \\

\rowcolor{categorycolor}
\multicolumn{12}{l}{\textit{\textbf{Causal-related}}} \\
& CI-GNN & 161K
  & 63.5\spm{14.0} & 18.8\spm{15.3} & 58.6\spm{8.8}
  & 51.2\spm{5.0}  & 55.0\spm{6.1}  & 50.0\spm{2.7}
  & 37.6\spm{4.2}  & 30.0\spm{3.6}  & 53.4\spm{5.8}  \\
& Contrasformer & 178K
  & 71.3\spm{1.4}  & 21.5\spm{15.2} & 52.8\spm{9.8}
  & 63.0\spm{1.3}  & 56.5\spm{5.3}  & 62.2\spm{2.1}
  & \underline{53.2\spm{4.6}}  & \underline{44.2\spm{3.8}}  & \underline{70.1\spm{5.5}}  \\
& CAL & 37K
  & 61.6\spm{6.0}  & 40.0\spm{17.5} & 56.1\spm{8.4}
  & 57.5\spm{2.7}  & 52.8\spm{3.9}  & 57.1\spm{2.6}
  & 45.6\spm{4.0}  & 34.5\spm{3.1}  & 61.9\spm{3.2}  \\

\midrule

\rowcolor{categorycolor}
\multicolumn{12}{l}{\textit{\textbf{Artemis (ours)}}} \\
& Artemis & 19K
  & \textbf{84.3\spm{3.5}}  & \textbf{73.5\spm{5.2}}  & \textbf{85.4\spm{5.6}}
  & \textbf{86.7\spm{3.8}}  & \textbf{85.5\spm{3.9}}  & \textbf{92.4\spm{3.1}}
  & \textbf{57.9\spm{3.4}}  & \textbf{45.8\spm{2.8}}  & \textbf{74.8\spm{3.5}}  \\
& {\scriptsize\textit{$\Delta$ vs.\ GCN}} & {\scriptsize\textit{+7K}}
  & \scriptsize\textit{+20.9} & \scriptsize\textit{+27.3} & \scriptsize\textit{+26.2}
  & \scriptsize\textit{+27.9} & \scriptsize\textit{+28.8} & \scriptsize\textit{+34.2}
  & \scriptsize\textit{+7.8}  & \scriptsize\textit{+5.4}  & \scriptsize\textit{+8.0}  \\
\bottomrule
\end{tabular}

\vspace{2pt}
{\scriptsize
Params counted with ADNI configuration ($N$=90, hidden\_dim=64).
The intervention adds only $\sim$7K parameters to any backbone.
}
\end{table*}

Table~\ref{tab:results} summarizes results across all three benchmarks.
Artemis outperforms every baseline on every dataset, with especially pronounced gains on the metrics most sensitive to class imbalance.
On ADNI, Artemis achieves 84.3\% accuracy, a $+5.0$\% absolute improvement over the best baseline BrainNetCNN (79.3\%), and a macro-F1 of 73.5\% compared with BrainNetCNN's 52.4\%, a $+21.1$ percentage-point gain that reflects far more balanced recall across the NC and MCI classes.
On HCP, Artemis reaches 86.7\% accuracy ($+6.2$\% over BrainNetCNN's 80.5\%) and 92.4\% AUC ($+8.7$\% over the best baseline AUC of 83.7\%).
On OASIS, Artemis improves accuracy by $+4.7$\% over Contrasformer (53.2\%) and delivers consistent gains across F1 and AUC.

Crucially, Artemis shares the same GCN backbone as the weakest baseline (vanilla GCN, $12$K parameters), while the only addition is the $7$K-parameter intervention module.
Adding this drop-in causal adjustment alone lifts accuracy by $+20.9$\%, $+27.9$\%, and $+7.8$\% on ADNI, HCP, and OASIS respectively (Table~\ref{tab:results}, $\Delta$ row), and macro-F1 by $+27.3$, $+28.8$, and $+5.4$ points.

It is worth mentioning that, the three causal baselines, CI-GNN, Contrasformer, and CAL consistently underperform the brain-specific architectures.
Their mechanisms (Granger causality, contrast graphs, and causal attention on structural subgraphs, respectively) do not account for the demographic confounders that dominate clinical brain imaging (e.g. CAL in particular, a general-purpose causal GNN, drops to $61.6\%$ accuracy on ADNI which even worse than vanilla GCN), reinforcing the motivation of Artemisß.
Meanwhile, the Artemis intervention module adds only ${\sim}7$K parameters on top of the GCN backbone (12K), a negligible overhead compared with Contrasformer (178K) or BrainNetCNN (882K), demonstrating that the gains originate from the causal intervention mechanism rather than model capacity.

\subsection{Ablation Study.}

\begin{table*}[t]
\centering
\caption{Ablation on intervention granularity.
         GCN backbone for all rows.
         FC-only / SC-only apply region-level gated intervention to one modality only.
         Single-stream concatenates $[f_i; s_i]$.
         Global shares a single confounder embedding across all ROIs.
         All rows use default hyperparameters.
}
\label{tab:ablation}
\footnotesize
\setlength{\tabcolsep}{3pt}
\begin{tabular}{l cc cc cc}
\toprule
& \multicolumn{2}{c}{\textbf{ADNI}} & \multicolumn{2}{c}{\textbf{HCP}} & \multicolumn{2}{c}{\textbf{OASIS}} \\
\cmidrule(lr){2-3}\cmidrule(lr){4-5}\cmidrule(lr){6-7}
\textbf{Intervention} & Acc & F1 & Acc & F1 & Acc & F1 \\
\midrule
Vanilla GCN
  & 63.4\spm{13.2} & 46.2\spm{11.6}
  & 58.8\spm{5.0}  & 56.7\spm{7.8}
  & 50.1\spm{5.5}  & 40.4\spm{13.6} \\
FC-only
  & 80.5\spm{6.2}  & 69.1\spm{9.3}
  & 82.8\spm{1.5}  & 81.9\spm{1.9}
  & 52.9\spm{3.2}  & 44.7\spm{3.2}  \\
SC-only
  & 80.2\spm{7.4}  & 67.2\spm{11.5}
  & 77.2\spm{2.9}  & 75.1\spm{2.9}
  & 52.9\spm{4.2}  & 45.2\spm{1.9}  \\
Single-stream
  & 81.1\spm{6.9}  & 70.2\spm{10.6}
  & 80.6\spm{1.8}  & 79.6\spm{1.6}
  & 52.4\spm{2.9}  & 43.2\spm{3.1}  \\
Global
  & 80.5\spm{5.7}  & 57.3\spm{28.9}
  & 78.4\spm{1.6}  & 75.9\spm{2.1}
  & 52.6\spm{3.5}  & 43.4\spm{4.3}  \\
\textbf{Artemis}
  & \textbf{84.3\spm{3.5}} & \textbf{73.5\spm{5.2}}
  & \textbf{86.7\spm{3.8}} & \textbf{85.5\spm{3.9}}
  & \textbf{57.9\spm{3.4}} & \textbf{45.8\spm{2.8}} \\
\bottomrule
\end{tabular}

\vspace{2pt}
\end{table*}

Table~\ref{tab:ablation} isolates each design choice under default hyperparameters.
The progression from no intervention to global and region-level is monotonic across all three benchmarks: global backdoor adjustment alone already lifts ADNI accuracy from $63.4$\% to $80.5$\%, and region-level differentiation adds another $+3.8$\% accuracy and $+16.2$ F1 points on ADNI, consistent with the view that heterogeneous per-ROI confounder effects cannot be captured by a single shared embedding; the large drop in F1 std (Global $28.9$ $\rightarrow$ Artemis $5.2$ on ADNI) further indicates that per-ROI parameterization acts as a regularizer.
Single-modality ablations (FC-only, SC-only) recover much of the gain but neither matches full Artemis, showing that the two modalities carry complementary confounding signal.
Finally, replacing the dual-stream gate with a single shared gate on the concatenated $[f_i;s_i]$ (Figure~\ref{fig:intervention} left) loses $3$-$6$\% accuracy across benchmarks (e.g., $80.6$\% vs.\ $86.7$\% on HCP), validating the modality-specific gating design.

\subsection{Parameter Sensitivity}

\begin{figure}[t]
    \centering
    \includegraphics[width=0.9\columnwidth]{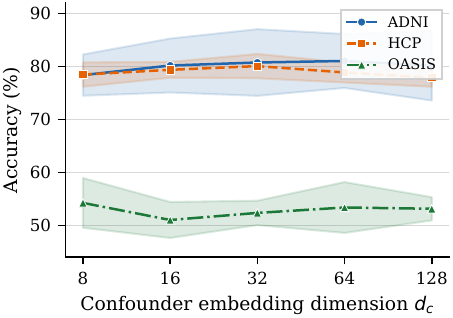}
    \caption{Sensitivity to confounder embedding dimension $d_c$. Accuracy remains stable ($\leq 3\%$ variation) across all three benchmarks.}
    \label{fig:dc_sensitivity}
\end{figure}

Figure~\ref{fig:dc_sensitivity} shows accuracy as a function of the confounder embedding dimension $d_c\!\in\!\{8,16,32,64,128\}$ under otherwise default hyperparameters.
Performance remains stable across all three benchmarks, with less than $3$\% variation in accuracy.
ADNI and HCP peak near $d_c\!=\!32$-$64$; larger values ($d_c\!=\!128$) offer no further gain and introduce mild overfitting.
This insensitivity confirms that the performance improvements stem from the causal intervention mechanism itself, not from additional model capacity.

\subsection{Interpretability Analysis}

We first visualize the pre-classifier embeddings with t-SNE on all three benchmarks (Figure~\ref{fig:tsne_support}): vanilla GCN embeddings conflate task labels with silhouette near zero, whereas Artemis yields clearly separable class structure across ADNI, HCP, and OASIS.

To examine \emph{where} the intervention acts, we compare per-ROI gate strength ($|g_i - 0.5|$, averaged over subjects and feature dimensions) with permutation importance (accuracy drop when shuffling ROI $i$) for all 90 ADNI ROIs (Figure~\ref{fig:quadrant}).
The two are largely uncorrelated (Spearman $\rho\!=\!{-0.06}$, $p\!=\!0.58$), confirming that the gate is driven by the centered confounder embedding, not by the task label, and therefore decoupled from classification importance by design.
The \emph{upper-right} quadrant (high gate, high importance) contains the left hippocampus and bilateral amygdala, regions central to Alzheimer's pathology and strongly modulated by aging~\cite{Fjell2014WhatIN}; Artemis applies targeted adjustment here, preserving the disease-relevant component while suppressing the age-driven shortcut.
The \emph{upper-left} quadrant (e.g., Frontal\_Sup\_R, ParaHippocampal\_R) contains age-sensitive but task-irrelevant regions whose high gate strength reflects suppression of a spurious backdoor path, while the \emph{lower-right} quadrant contains clean biomarkers that require no adjustment.
This pattern validates that Artemis calibrates precisely the regions where demographic and disease effects are entangled.

\begin{figure}[!t]
    \centering
    \includegraphics[width=\columnwidth]{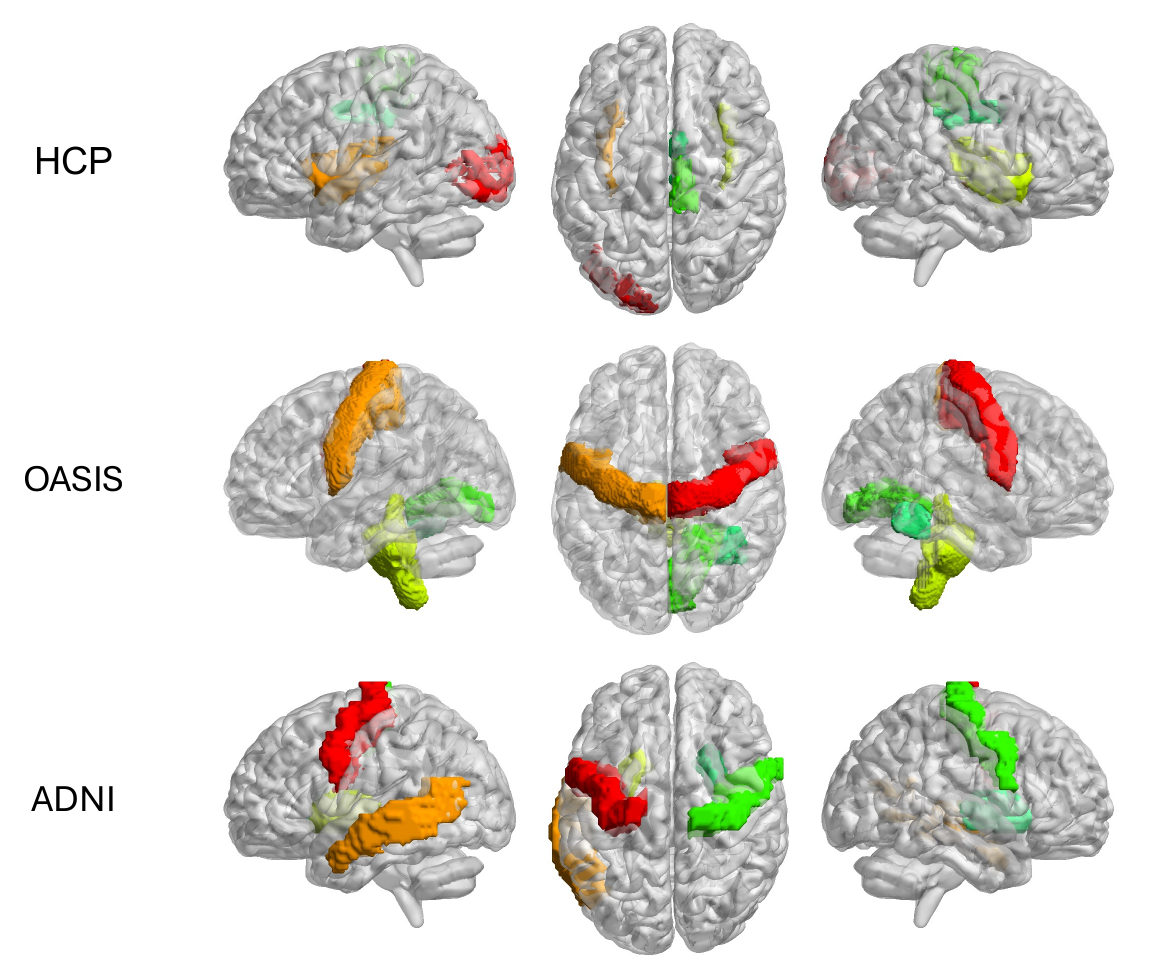}
    \caption{Integrated-Gradient attribution of the top-5 ROIs for HCP (top), OASIS (middle), and ADNI (bottom). ROIs driving the task decision after \textsc{Artemis} intervention.}
    \label{fig:brain_roi_inter}
\end{figure}

To locate the regions that drive the task decision \emph{after} the intervention, we compute Integrated Gradients (IG, 50 steps, zero baseline) of the predicted class w.r.t.\ the per-ROI input features for every subject, and project the top-5 ROIs per benchmark onto the cortical surface (Figure~\ref{fig:brain_roi_inter}).
On ADNI, IG highlights bilateral precentral gyrus, left middle temporal gyrus, and bilateral putamen, consistent with motor-cortex thinning and striatal tauopathy reported in advancing AD~\cite{whitwell2008rates}, while hippocampus and amygdala (ranked 54-88) are deliberately down-weighted because their variance is already explained by the demographic gate.
On HCP, the top-5 concentrates on bilateral insula, posterior cingulate, lateral occipital, and paracentral cortex, regions with the strongest documented structural sex dimorphism in the connectome~\cite{ingalhalikar2014sex,Ritchie2017SexDI}.
On OASIS, bilateral precentral gyrus, lingual gyrus, and temporo-occipital fusiform emerge as most informative, matching the posterior-dominant atrophy pattern observed in dementia cohorts~\cite{lehmann2010atrophy}.
The contrast between IG (task-driven attribution) and gate strength (confounder-driven modulation) on ADNI is complementary: the gate suppresses demographically confounded directions in medial-temporal regions, allowing the GCN to ground its decision in the remaining, less-confounded motor-subcortical signal.

\section{Conclusion.}
\label{sec:conclusion}
We presented Artemis, a region-level causal intervention framework that addresses demographic confounding in multimodal brain-network analysis.
By mapping observed demographics to per-ROI confounder embeddings via a shared MLP with learnable region tokens, centering each embedding against an EMA population bank, and applying a learned sigmoid gate to both functional and structural features, Artemis performs backdoor adjustment at each brain region independently.
The entire intervention adds only a few thousand parameters and plugs into any GNN backbone without architectural modification.
Experiments on ADNI, HCP, and OASIS demonstrate consistent improvements over ten representative baselines, and ROI-level analyses confirm that the intervention targets confounder-sensitive regions while remaining decoupled from task-relevant signals.

\subsection{Limitations and Future Work}
Artemis models only \emph{observed} demographic confounders (age, sex, education) and evaluates on cross-sectional cohorts, unmeasured confounders (e.g., socioeconomic status, medication) and longitudinal dynamics are not yet considered.
We plan to (i)~extend Artemis to implicit confounders via a post-backbone adversarial, (ii)~apply it to regression targets (e.g., MMSE, depression/anxiety scores), and (iii)~scale to multi-site cohorts for further scanner-effected demographic confoundings.

\bibliographystyle{siamplain}
\bibliography{main}
\end{document}